\documentclass[11pt]{article}

\usepackage{amsthm,amssymb, epsfig, epstopdf, tikz, tikz-cd, tcolorbox, subcaption, setspace, graphicx, hyperref, enumitem, mathrsfs, bbm, mathtools,color,stmaryrd, comment, fancyheadings, mathpazo, euler, subcaption,cleveref}
\theoremstyle{definition}

\usepackage[margin=.9in]{geometry}
\newtheorem{remark}{Remark}[section]

\newtheorem{theorem}{Theorem}
\newtheorem{definition}{Definition}[section]

\newtheorem{lemma}{Lemma}[section]

\begin{document}
\title{Taylor Learning}
\author{James Schmidt\footnote{Department of Applied Mathematics and Statistics, email: aschmi40@jhu.edu}\\
 Johns Hopkins University}
 
\date{\today}
\maketitle
\abstract{Empirical risk minimization stands behind most optimization in supervised machine learning. Under this scheme, labeled data is used to approximate an expected cost (risk), and a learning algorithm updates model-defining parameters in search of an empirical risk minimizer, with the aim of  thereby approximately minimizing expected cost. Parameter update  is often done by some sort of gradient descent. In this paper, we introduce a learning algorithm to construct models for real analytic functions using neither gradient descent nor empirical risk minimization. Observing that such functions are defined by local information, we situate familiar Taylor approximation methods in the context of sampling data from a distribution, and prove a nonuniform learning result.}
\newcommand{\nat}[6][large]{%
  \begin{tikzcd}[ampersand replacement = \&, column sep=#1]
    #2\ar[bend left=40,""{name=U}]{r}{#4}\ar[bend right=40,',""{name=D}]{r}{#5}\& #3
          \ar[shorten <=10pt,shorten >=10pt,Rightarrow,from=U,to=D]{d}{~#6}
    \end{tikzcd}
}
\newcommand{\invamalg}{\mathbin{\text{\rotatebox[origin=c]{180}{$\amalg$}}}}

\newcommand{\dCrl}[0]{\mathfrak{dCrl}}
\newcommand{\ytil}{\tilde{y}}
\newcommand{\defeq}{\vcentcolon=}
\newcommand{\dee}{\partial}
\newcommand{\lb}{\{}
\newcommand{\rb}{\}}
\newcommand{\R}{\mathbb{R}}
\newcommand{\C}{\mathbb{C}}
\newcommand{\Q}{\mathbb{Q}}
\newcommand{\N}{\mathbb{N}}
\newcommand{\el}{\mathcal{L}}
\newcommand{\pdiv}[2]{\frac{\partial{#1}}{\partial{#2}}}
\newcommand{\discatp}{\displaystyle\bigsqcap}
\newcommand{\discats}{\displaystyle\bigsqcup}

\newcommand{\uZ}{\underline{\mathbb{Z}}}
\newcommand{\uF}[1]{\underline{\mathbb{F}}}
\newcommand{\one}{\mathbb{1}}
\newcommand{\two}{\mathbb{2}}

\newcommand{\dis}{\displaystyle}
\newcommand{\disp}{\displaystyle\prod}
\newcommand{\disu}{\displaystyle\bigcup}
\newcommand{\disi}{\displaystyle\bigcap}
\newcommand{\diss}{\displaystyle\sum}
\newcommand{\disg}{\displaystyle\int}
\newcommand{\disl}{\displaystyle\lim}
\newcommand{\dislim}{\displaystyle\lim}
\newcommand{\disliminf}{\displaystyle\liminf}
\newcommand{\dislimsup}{\displaystyle\limsup}
\newcommand{\disbop}{\displaystyle\bigotimes}
\newcommand{\disbos}{\displaystyle\bigoplus}
\newcommand{\dissup}{\displaystyle\sup}
\newcommand{\disinf}{\displaystyle\inf}
\newcommand{\dismax}{\displaystyle\max}
\newcommand{\dismin}{\displaystyle\min}
\newcommand{\dirlim}[1]{\displaystyle\varinjlim_{#1}}
\newcommand{\indlim}[1]{\displaystyle\varprojlim_{#1}}
\newcommand{\discatlim}{\indlim}
\newcommand{\discatcolim}{\dirlim}
\newcommand{\catcolim}{\mbox{colim}}
\newcommand{\catlim}{\mbox{lim}}

\newcommand{\colgray}[1]{\color{gray}{#1}\color{black}}

\newcommand{\sfM}{\sF{M}}

\newcommand{\forget}[2]{\Ub^{#1}_{#2}}

\newcommand\righttwoarrow{%
        \mathrel{\vcenter{\mathsurround0pt
                \ialign{##\crcr
                        \noalign{\nointerlineskip}$\rightarrow$\crcr
                        \noalign{\nointerlineskip}$\rightarrow$\crcr
                }%
        }}%
}

\newcommand{\Z}{\mathbb{Z}}
\newcommand{\Ab}[0]{\mathbb{A}}
\newcommand{\Bb}[0]{\mathbb{B}}
\newcommand{\Cb}[0]{\mathbb{C}}
\newcommand{\Db}[0]{\mathbb{D}}
\newcommand{\Eb}[0]{\mathbb{E}}
\newcommand{\Fb}[0]{\mathbb{F}}
\newcommand{\Gb}[0]{\mathbb{G}}
\newcommand{\Hb}[0]{\mathbb{H}}
\newcommand{\Ib}[0]{\mathbb{I}}
\newcommand{\Jb}[0]{\mathbb{J}}
\newcommand{\Kb}[0]{\mathbb{K}}
\newcommand{\Lb}[0]{\mathbb{L}}
\newcommand{\Mb}[0]{\mathbb{M}}
\newcommand{\Nb}[0]{\mathbb{N}}
\newcommand{\Ob}[0]{\mathbb{O}}
\newcommand{\Pb}[0]{\mathbb{P}}
\newcommand{\Qb}[0]{\mathbb{Q}}
\newcommand{\Rb}[0]{\mathbb{R}}
\newcommand{\Sb}[0]{\mathbb{S}}
\newcommand{\Tb}[0]{\mathbb{T}}
\newcommand{\Ub}[0]{\mathbb{U}}
\newcommand{\Vb}[0]{\mathbb{V}}
\newcommand{\Wb}[0]{\mathbb{W}}
\newcommand{\Xb}[0]{\mathbb{X}}
\newcommand{\Yb}[0]{\mathcal{Y}}
\newcommand{\Zb}[0]{\mathbb{Z}}

\newcommand{\sF}[1]{\mathsf{#1}}

\newcommand{\sC}[1]{\mathscr{#1}}

\newcommand{\mC}[1]{\mathcal{#1}}

\newcommand{\mB}[1]{\mathbb{#1}}

\newcommand{\mF}[1]{\mathfrak{#1}}

\newcommand{\Bc}[0]{\mathcal{B}}
\newcommand{\Cc}[0]{\mathcal{C}}
\newcommand{\Dc}[0]{\mathcal{D}}
\newcommand{\Ec}[0]{\mathcal{E}}
\newcommand{\Fc}[0]{\mathcal{F}}
\newcommand{\Gc}[0]{\mathcal{G}}
\newcommand{\Hc}[0]{\mathcal{H}}
\newcommand{\Ic}[0]{\mathcal{I}}
\newcommand{\Jc}[0]{\mathcal{J}}
\newcommand{\Kc}[0]{\mathcal{K}}
\newcommand{\Lc}[0]{\mathcal{L}}
\newcommand{\Mc}[0]{\mathcal{M}}
\newcommand{\Nc}[0]{\mathcal{N}}
\newcommand{\Oc}[0]{\mathcal{O}}
\newcommand{\Pc}[0]{\mathcal{P}}
\newcommand{\Qc}[0]{\mathcal{Q}}
\newcommand{\Rc}[0]{\mathcal{R}}
\newcommand{\Sc}[0]{\mathcal{S}}
\newcommand{\Tc}[0]{\mathcal{T}}
\newcommand{\Uc}[0]{\mathcal{U}}
\newcommand{\Vc}[0]{\mathcal{V}}
\newcommand{\Wc}[0]{\mathcal{W}}
\newcommand{\Xc}[0]{\mathcal{X}}
\newcommand{\Yc}[0]{\mathcal{Y}}
\newcommand{\Zc}[0]{\mathcal{Z}}

\newcommand{\aca}[0]{\mathcal{a}}
\newcommand{\bca}[0]{\mathcal{b}}
\newcommand{\cca}[0]{\mathcal{c}}
\newcommand{\dca}[0]{\mathcal{d}}
\newcommand{\eca}[0]{\mathcal{e}}
\newcommand{\fca}[0]{\mathcal{f}}
\newcommand{\gca}[0]{\mathcal{g}}
\newcommand{\hca}[0]{\mathcal{h}}
\newcommand{\ica}[0]{\mathcal{i}}
\newcommand{\jca}[0]{\mathcal{j}}
\newcommand{\kca}[0]{\mathcal{k}}
\newcommand{\lca}[0]{\mathcal{l}}
\newcommand{\mca}[0]{\mathcal{m}}
\newcommand{\nca}[0]{\mathcal{n}}
\newcommand{\oca}[0]{\mathcal{o}}
\newcommand{\pca}[0]{\mathcal{p}}
\newcommand{\qca}[0]{\mathcal{q}}
\newcommand{\rca}[0]{\mathcal{r}}
\newcommand{\sca}[0]{\mathcal{s}}
\newcommand{\tca}[0]{\mathcal{t}}
\newcommand{\uca}[0]{\mathcal{u}}
\newcommand{\vca}[0]{\mathcal{v}}
\newcommand{\wca}[0]{\mathcal{w}}
\newcommand{\xca}[0]{\mathcal{x}}
\newcommand{\yca}[0]{\mathcal{y}}
\newcommand{\zca}[0]{\mathcal{z}}

\newcommand{\Af}[0]{\mathfrak{A}}
\newcommand{\Bf}[0]{\mathfrak{B}}
\newcommand{\Cf}[0]{\mathfrak{C}}
\newcommand{\Df}[0]{\mathfrak{D}}
\newcommand{\Ef}[0]{\mathfrak{E}}
\newcommand{\Ff}[0]{\mathfrak{F}}
\newcommand{\Gf}[0]{\mathfrak{G}}
\newcommand{\Hf}[0]{\mathfrak{H}}
\newcommand{\If}[0]{\mathfrak{I}}
\newcommand{\Jf}[0]{\mathfrak{J}}
\newcommand{\Kf}[0]{\mathfrak{K}}
\newcommand{\Lf}[0]{\mathfrak{L}}
\newcommand{\Mf}[0]{\mathfrak{M}}
\newcommand{\Nf}[0]{\mathfrak{N}}
\newcommand{\Of}[0]{\mathfrak{O}}
\newcommand{\Pf}[0]{\mathfrak{P}}
\newcommand{\Qf}[0]{\mathfrak{Q}}
\newcommand{\Rf}[0]{\mathfrak{R}}
\newcommand{\Sf}[0]{\mathfrak{S}}
\newcommand{\Tf}[0]{\mathfrak{T}}
\newcommand{\Uf}[0]{\mathfrak{U}}
\newcommand{\Vf}[0]{\mathfrak{V}}
\newcommand{\Wf}[0]{\mathfrak{W}}
\newcommand{\Xf}[0]{\mathfrak{X}}
\newcommand{\Yf}[0]{\mathfrak{Y}}
\newcommand{\Zf}[0]{\mathfrak{Z}}

\newcommand{\af}[0]{\mathfrak{a}}
\newcommand{\bff}[0]{\mathfrak{b}}
\newcommand{\cf}[0]{\mathfrak{c}}
\newcommand{\dff}[0]{\mathfrak{d}}
\newcommand{\ef}[0]{\mathfrak{e}}
\newcommand{\ff}[0]{\mathfrak{f}}
\newcommand{\gf}[0]{\mathfrak{g}}
\newcommand{\hf}[0]{\mathfrak{h}}
\newcommand{\ifrak}{\mathfrak{i}}
\newcommand{\jf}[0]{\mathfrak{j}}
\newcommand{\kf}[0]{\mathfrak{k}}
\newcommand{\lf}[0]{\mathfrak{l}}
\newcommand{\mf}[0]{\mathfrak{m}}
\newcommand{\nf}[0]{\mathfrak{n}}
\newcommand{\of}[0]{\mathfrak{o}}
\newcommand{\pf}[0]{\mathfrak{p}}
\newcommand{\qf}[0]{\mathfrak{q}}
\newcommand{\rf}[0]{\mathfrak{r}}
\renewcommand{\sf}[0]{\mathfrak{s}}
\newcommand{\tf}[0]{\mathfrak{t}}
\newcommand{\uf}[0]{\mathfrak{u}}
\newcommand{\vf}[0]{\mathfrak{v}}
\newcommand{\wf}[0]{\mathfrak{w}}
\newcommand{\xf}[0]{\mathfrak{x}}
\newcommand{\yf}[0]{\mathfrak{y}}
\newcommand{\zf}[0]{\mathfrak{z}}
\newcommand{\cdX}{\mathfrak{cdX}}
\newcommand{\cdCrl}{\mathfrak{cdCrl}}

\newcommand{\scA}[0]{\mathscr{A}}
\newcommand{\scB}[0]{\mathscr{B}}
\newcommand{\scC}[0]{\mathscr{C}}
\newcommand{\scD}[0]{\mathscr{D}}
\newcommand{\scE}[0]{\mathscr{E}}
\newcommand{\scF}[0]{\mathscr{F}}
\newcommand{\scG}[0]{\mathscr{G}}
\newcommand{\scH}[0]{\mathscr{H}}
\newcommand{\scI}[0]{\mathscr{I}}
\newcommand{\scJ}[0]{\mathscr{J}}
\newcommand{\scK}[0]{\mathscr{K}}
\newcommand{\scL}[0]{\mathscr{L}}
\newcommand{\scM}[0]{\mathscr{M}}
\newcommand{\scN}[0]{\mathscr{N}}
\newcommand{\scO}[0]{\mathscr{O}}
\newcommand{\scP}[0]{\mathscr{P}}
\newcommand{\scQ}[0]{\mathscr{Q}}
\newcommand{\scR}[0]{\mathscr{R}}
\newcommand{\scS}[0]{\mathscr{S}}
\newcommand{\scT}[0]{\mathscr{T}}
\newcommand{\scU}[0]{\mathscr{U}}
\newcommand{\scV}[0]{\mathscr{V}}
\newcommand{\scW}[0]{\mathscr{W}}
\newcommand{\scX}[0]{\mathscr{X}}
\newcommand{\scY}[0]{\mathscr{Y}}
\newcommand{\scZ}[0]{\mathscr{Z}}

\newcommand{\fA}[0]{\mathsf{A}}
\newcommand{\fB}[0]{\mathsf{B}}
\newcommand{\fC}[0]{\mathsf{C}}
\newcommand{\fD}[0]{\mathsf{D}}
\newcommand{\fE}[0]{\mathsf{E}}
\newcommand{\fG}[0]{\mathsf{G}}
\newcommand{\fH}[0]{\mathsf{H}}
\newcommand{\fI}[0]{\mathsf{I}}
\newcommand{\fJ}[0]{\mathsf{J}}
\newcommand{\fK}[0]{\mathsf{K}}
\newcommand{\fL}[0]{\mathsf{L}}
\newcommand{\fM}[0]{\mathsf{M}}
\newcommand{\fN}[0]{\mathsf{N}}
\newcommand{\fO}[0]{\mathsf{O}}
\newcommand{\fP}[0]{\mathsf{P}}
\newcommand{\fQ}[0]{\mathsf{Q}}
\newcommand{\fR}[0]{\mathsf{R}}
\newcommand{\fS}[0]{\mathsf{S}}
\newcommand{\fT}[0]{\mathsf{T}}
\newcommand{\fU}[0]{\mathsf{U}}
\newcommand{\fV}[0]{\mathsf{V}}
\newcommand{\fW}[0]{\mathsf{W}}
\newcommand{\fX}[0]{\mathsf{X}}
\newcommand{\fY}[0]{\mathsf{Y}}
\newcommand{\fZ}[0]{\mathsf{Z}}

\newcommand{\fa }[0]{\mathsf{a}}
\newcommand{\fb }[0]{\mathsf{b}}
\newcommand{\fc }[0]{\mathsf{c}}
\newcommand{\fd}[0]{\mathsf{d}}
\newcommand{\fe}[0]{\mathsf{e}}
\newcommand{\fg}[0]{\mathsf{g}}
\newcommand{\fh}[0]{\mathsf{h}}
\newcommand{\fj}[0]{\mathsf{j}}
\newcommand{\fk}[0]{\mathsf{k}}
\newcommand{\fl }[0]{\mathsf{l}}
\newcommand{\fm }[0]{\mathsf{m}}
\newcommand{\fn }[0]{\mathsf{n}}
\newcommand{\fo }[0]{\mathsf{o}}
\newcommand{\fp}[0]{\mathsf{p}}
\newcommand{\fq}[0]{\mathsf{q}}
\newcommand{\fr}[0]{\mathsf{r}}
\newcommand{\fs}[0]{\mathsf{s}}
\newcommand{\ft }[0]{\mathsf{t}}
\newcommand{\fu }[0]{\mathsf{u}}
\newcommand{\fv }[0]{\mathsf{v}}
\newcommand{\fw}[0]{\mathsf{w}}
\newcommand{\fx}[0]{\mathsf{x}}
\newcommand{\fy}[0]{\mathsf{y}}
\newcommand{\fz}[0]{\mathsf{z}}

\newcommand{\dA}[0]{\dot{A}}
\newcommand{\dB}[0]{\dot{B}}
\newcommand{\dC}[0]{\dot{C}}
\newcommand{\dD}[0]{\dot{D}}
\newcommand{\dE}[0]{\dot{E}}
\newcommand{\dF}[0]{\dot{F}}
\newcommand{\dG}[0]{\dot{G}}
\newcommand{\dH}[0]{\dot{H}}
\newcommand{\dI}[0]{\dot{I}}
\newcommand{\dJ}[0]{\dot{J}}
\newcommand{\dK}[0]{\dot{K}}
\newcommand{\dL}[0]{\dot{L}}
\newcommand{\dM}[0]{\dot{M}}
\newcommand{\dN}[0]{\dot{N}}
\newcommand{\dO}[0]{\dot{O}}
\newcommand{\dP}[0]{\dot{P}}
\newcommand{\dQ}[0]{\dot{Q}}
\newcommand{\dR}[0]{\dot{R}}
\newcommand{\dS}[0]{\dot{S}}
\newcommand{\dT}[0]{\dot{T}}
\newcommand{\dU}[0]{\dot{U}}
\newcommand{\dV}[0]{\dot{V}}
\newcommand{\dW}[0]{\dot{W}}
\newcommand{\dX}[0]{\dot{X}}
\newcommand{\dY}[0]{\dot{Y}}
\newcommand{\dZ}[0]{\dot{Z}}

\newcommand{\da}[0]{\dot{a}}
\newcommand{\db}[0]{\dot{b}}
\newcommand{\dc}[0]{\dot{c}}
\newcommand{\dd}[0]{\dot{d}}
\newcommand{\de}[0]{\dot{e}}
\newcommand{\df}[0]{\dot{f}}
\newcommand{\dg}[0]{\dot{g}}
\renewcommand{\dh}[0]{\dot{h}}
\newcommand{\di}[0]{\dot{i}}
\renewcommand{\dj}[0]{\dot{j}}
\newcommand{\dk}[0]{\dot{k}}
\newcommand{\dl}[0]{\dot{l}}
\newcommand{\dm}[0]{\dot{m}}
\newcommand{\dn}[0]{\dot{n}}
\newcommand{\dq}[0]{\dot{q}}
\newcommand{\dr}[0]{\dot{r}}
\newcommand{\ds}[0]{\dot{s}}
\newcommand{\dt}[0]{\dot{t}}
\newcommand{\du}[0]{\dot{u}}
\newcommand{\dv}[0]{\dot{v}}
\newcommand{\dw}[0]{\dot{w}}
\newcommand{\dx}[0]{\dot{x}}
\newcommand{\dy}[0]{\dot{y}}
\newcommand{\dz}[0]{\dot{z}}

\newcommand{\oA}[0]{\overline{A}}
\newcommand{\oB}[0]{\overline{B}}
\newcommand{\oC}[0]{\overline{C}}
\newcommand{\oD}[0]{\overline{D}}
\newcommand{\oE}[0]{\overline{E}}
\newcommand{\oF}[0]{\overline{F}}
\newcommand{\oG}[0]{\overline{G}}
\newcommand{\oH}[0]{\overline{H}}
\newcommand{\oI}[0]{\overline{I}}
\newcommand{\oJ}[0]{\overline{J}}
\newcommand{\oK}[0]{\overline{K}}
\newcommand{\oL}[0]{\overline{L}}
\newcommand{\oM}[0]{\overline{M}}
\newcommand{\oN}[0]{\overline{N}}
\newcommand{\oO}[0]{\overline{O}}
\newcommand{\oP}[0]{\overline{P}}
\newcommand{\oQ}[0]{\overline{Q}}
\newcommand{\oR}[0]{\overline{R}}
\newcommand{\oS}[0]{\overline{S}}
\newcommand{\oT}[0]{\overline{T}}
\newcommand{\oU}[0]{\overline{U}}
\newcommand{\oV}[0]{\overline{V}}
\newcommand{\oW}[0]{\overline{W}}
\newcommand{\oX}[0]{\overline{X}}
\newcommand{\oY}[0]{\overline{Y}}
\newcommand{\oZ}[0]{\overline{Z}}

\newcommand{\oa}[0]{\overline{a}}
\newcommand{\ob}[0]{\overline{b}}
\newcommand{\oc}[0]{\overline{c}}
\newcommand{\od}[0]{\overline{d}}
\renewcommand{\oe}[0]{\overline{e}}
\newcommand{\og}[0]{\overline{g}}
\newcommand{\oh}[0]{\overline{h}}
\newcommand{\oi}[0]{\overline{i}}
\newcommand{\oj}[0]{\overline{j}}
\newcommand{\ok}[0]{\overline{k}}
\newcommand{\ol}[0]{\overline{l}}
\newcommand{\om}[0]{\overline{m}}
\newcommand{\on}[0]{\overline{n}}
\newcommand{\oo}[0]{\overline{o}}
\newcommand{\op}[0]{\overline{p}}
\newcommand{\oq}[0]{\overline{q}}
\newcommand{\os}[0]{\overline{s}}
\newcommand{\ot}[0]{\overline{t}}
\newcommand{\ou}[0]{\overline{u}}
\newcommand{\ov}[0]{\overline{v}}
\newcommand{\ow}[0]{\overline{w}}
\newcommand{\ox}[0]{\overline{x}}
\newcommand{\oy}[0]{\overline{y}}
\newcommand{\oz}[0]{\overline{z}}

\renewcommand{\a}{\alpha}
\renewcommand{\b}{\beta}
\renewcommand{\d}{\delta}
\newcommand{\e}{\varepsilon}
\newcommand{\f}{\phi}
\newcommand{\g}{\gamma}
\newcommand{\h}{\eta}
\renewcommand{\i}{\iota}
\renewcommand{\k}{\kappa}
\renewcommand{\l}{\lambda}
\newcommand{\m}{\mu}
\newcommand{\n}{\nu}
\newcommand{\p}{\pi}
\newcommand{\ph}{\varphi}
\newcommand{\ps}{\psi}
\newcommand{\q}{\xi}
\renewcommand{\r}{\rho}
\newcommand{\s}{\sigma}
\renewcommand{\t}{\tau}
\renewcommand{\v}{\upsilon}
\newcommand{\x}{\chi}
\newcommand{\z}{\zeta}
\newcommand{\G}{\Gamma}

\newcommand{\aarb}[0]{\<a>}
\newcommand{\barb}[0]{\<b>}
\newcommand{\carb}[0]{\<c>}
\newcommand{\darb}[0]{\<d>}
\newcommand{\earb}[0]{\<e>}
\newcommand{\farb}[0]{\<f>}
\newcommand{\garb}[0]{\<g>}
\newcommand{\harb}[0]{\<h>}
\newcommand{\iarb}[0]{\<i>}
\newcommand{\jarb}[0]{\<j>}
\newcommand{\karb}[0]{\<k>}
\newcommand{\larb}[0]{\<l>}
\newcommand{\marb}[0]{\<m>}
\newcommand{\narb}[0]{\<n>}
\newcommand{\oarb}[0]{\<o>}
\newcommand{\parb}[0]{\<p>}
\newcommand{\qarb}[0]{\<q>}
\newcommand{\rarb}[0]{\<r>}
\newcommand{\sarb}[0]{\<s>}
\newcommand{\tarb}[0]{\<t>}
\newcommand{\uarb}[0]{\<u>}
\newcommand{\varb}[0]{\<v>}
\newcommand{\warb}[0]{\<w>}
\newcommand{\xarb}[0]{\<x>}
\newcommand{\yarb}[0]{\<y>}
\newcommand{\zarb}[0]{\<z>}

\newcommand{\hA}[0]{\hat{A}}
\newcommand{\hB}[0]{\hat{B}}
\newcommand{\hC}[0]{\hat{C}}
\newcommand{\hD}[0]{\hat{D}}
\newcommand{\hE}[0]{\hat{E}}
\newcommand{\hF}[0]{\hat{F}}
\newcommand{\hG}[0]{\hat{G}}
\newcommand{\hH}[0]{\hat{H}}
\newcommand{\hI}[0]{\hat{I}}
\newcommand{\hJ}[0]{\hat{J}}
\newcommand{\hK}[0]{\hat{K}}
\newcommand{\hL}[0]{\hat{L}}
\newcommand{\hM}[0]{\hat{M}}
\newcommand{\hN}[0]{\hat{N}}
\newcommand{\hO}[0]{\hat{O}}
\newcommand{\hP}[0]{\hat{P}}
\newcommand{\hQ}[0]{\hat{Q}}
\newcommand{\hR}[0]{\hat{R}}
\newcommand{\hS}[0]{\hat{S}}
\newcommand{\hT}[0]{\hat{T}}
\newcommand{\hU}[0]{\hat{U}}
\newcommand{\hV}[0]{\hat{V}}
\newcommand{\hW}[0]{\hat{W}}
\newcommand{\hX}[0]{\hat{X}}
\newcommand{\hY}[0]{\hat{Y}}
\newcommand{\hZ}[0]{\hat{Z}}

\newcommand{\ha}[0]{\hat{a}}
\newcommand{\hb}[0]{\hat{b}}
\newcommand{\hc}[0]{\hat{c}}
\newcommand{\hd}[0]{\hat{d}}
\newcommand{\he}[0]{\hat{e}}
\newcommand{\hg}[0]{\hat{g}}
\newcommand{\hh}[0]{\hat{h}}
\newcommand{\hi}[0]{\hat{i}}
\newcommand{\hj}[0]{\hat{j}}
\newcommand{\hk}[0]{\hat{k}}
\newcommand{\hl}[0]{\hat{l}}
\newcommand{\hm}[0]{\hat{m}}
\newcommand{\hn}[0]{\hat{n}}
\newcommand{\ho}[0]{\hat{o}}
\newcommand{\hp}[0]{\hat{p}}
\newcommand{\hq}[0]{\hat{q}}
\newcommand{\hr}[0]{\hat{r}}
\newcommand{\hs}[0]{\hat{s}}
\newcommand{\hu}[0]{\hat{u}}
\newcommand{\hv}[0]{\hat{v}}
\newcommand{\hw}[0]{\hat{w}}
\newcommand{\hx}[0]{\hat{x}}
\newcommand{\hy}[0]{\hat{y}}
\newcommand{\hz}[0]{\hat{z}}

\newcommand{\hyph}[2]{\Fc_{#1}:#2\rightarrow \sF{RelMan^c}}
\newcommand{\repsys}{(\Fc_{N^\scT}:\N\rightarrow\sF{RelMan^c},\frac{d}{dt})}
\newcommand{\truerep}[1]{\left(\Fc_{N^\scT}{#1}:\N\rightarrow\sF{RelMan^c},\frac{D}{Dt}\right)}

\section{Introduction}

Empirical risk minimization forms the backbone of supervised machine learning: a finite labeled data set is used to search a function space for a model which fits both the data and the distribution generating it. Intuition for why fitting empirical risk ought to fit expectation may crudely derive from Law of Large Numbers reasoning, but generally uniform guarantees, such as PAC learnability, provide rigorous grounds for its use.  In the absence of uniform guarantees and presence of particularized knowledge of  data, one may consider other schemes for constructing a model. 

We offer one, using insight from calculus that local information for a class of real analytic functions provides global information. Together with the definition of derivative as limit, we observe that sampled function evaluations may provide arbitrarily fine approximations of derivatives to arbitrarily high order, and sampling enough is guaranteed to produce samples close enough to a point of interest. Thus we turn Taylor polynomial principles into a learning algorithm and show that under some conditions on the measure generating data, this procedure is guaranteed to produce a well-approximating function in probability.

The result presents a preliminary effort to merge sampling in numerical analysis with probabilistic sampling. More fundamentally, it makes use of  \textit{nonuniform}  learnability notions (\cite{schmidtOverfitting}, akin to \cite[Ch.\ 7 ]{uml}), namely  precision guarantees with confidence whose sample complexity may not be independent of the data-generating distribution. While weaker than e.g.\ PAC learnability, this version of learnability requires careful attention to peculiarities of both the hypothesis class and the measure, and  we anticipate that other concrete and interesting examples abound showing its use. 

In  preliminary detail, we will show that a real analytic function may be well approximated by taking  finitely many function evaluations to form an approximation of a sufficiently high degree Taylor polynomial. Where this result diverges from ordinary Taylor approximation statements is that 1.\ derivatives $f(p), f'(p),\ldots,f^{(n)}(p)$ are to be \textit{approximated} via  2.\ sampled data which comes from a distribution. In bounding the expected error, we shall decompose the integral into two components, the tail of the integral and the integral on a compact domain. We argue that under mild assumptions on the distribution, the tail may be made arbitrarily small. On the compact domain, we show that with high probability, the derivatives $f^{(j)}(p)$ for $j=0,\ldots,n$ may be approximated well enough provided enough data is sampled.

In \cref{sec:pacreg}, we review relevant notions from calculus and learning theory. We start in \cref{subsection:taylor_series} with Taylor series and convergence, and by extension Taylor polynomials and their approximation properties. In \cref{subsection:P-Learning} we then review   a variant of PAC learnability (\cite{schmidtOverfitting}), which removes assumptions and guarantees of uniformity with respect to measure. We apply this modified notion to regression tasks, where  both models and costs may be unbounded. The background  prepares for \cref{sec:taylor_learning} where we   present and prove the  main theorem which says that any analytic function may be learned using the class of polynomials  provided noiseless data is generated from a distribution without fat tails. 

\section{Background}\label{sec:pacreg}\label{sec:background}
\subsection{Taylor Series and Analytic Functions}\label{subsection:taylor_series}

Taylor series and polynomials contain a kernel of  calculus (\cite{apostol}, \cite{spivakcalculus}), and distill much of integration and differentiation to knowledge of such on monomials: $$ \begin{array}{lll} \disg x^n dx = \frac{x^{n+1}}{n+1} & \mbox{and} & \frac{d}{dx} x^n = nx^{n-1}.\end{array}$$ Particularly of use is that monomials $x^n$---and by extension, polynomials generally---are computable. Moreover, Taylor computation  well-captures the very concept of derivative. While derivatives are often  introduced  through a  limiting procedure on secant approximations for a function's slope, a better perspective is that they induce   a line of best fit at a point (\cite{follandcalculus}): \begin{equation}\label{eq:taylor1} f(h) = f(0)+ f'(0)h + o(h),\end{equation} where little-oh  is defined implicitly by satisfaction of the limit $\disl_{h\rightarrow0}\frac{o(h)}{h} = 0$; the little-oh term captures how quickly the error term dissipates.   Higher order Taylor polynomials accentuate this point: $$f(h) = f(0) + f'(0)h + \frac{1}{2}f''(0) h^2 + o(h^2)$$ is a quadratic of best fit at $0$, with error term $o(h^2)$ that approaches zero much faster: $$\disl_{h\rightarrow0}\frac{o(h^2)}{h^2} = 0.$$
\begin{figure}[h!]
\centerline{\includegraphics[scale = 1.5]{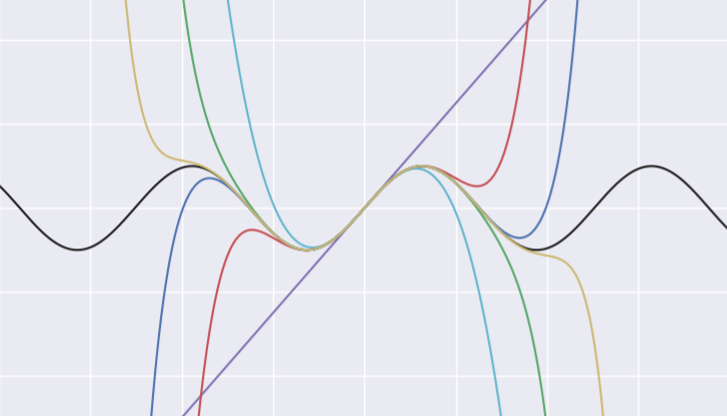}}\caption{Taylor polynomials for $\sin$}\label{figure:taylorApprox}	
\end{figure}

Speed of convergence as $h\rightarrow 0$ is  one  way to think about higher order derivatives: they define the best fit polynomial of degree $n$ at a point whose fit is quantified at that point by error term $o(h^n)$. Yet another perspective  is that the Taylor polynomial  approximates a function \textit{away} from $h\approx0$: consider plots in \cref{figure:taylorApprox} which indicate  approximations far from point of expansion for various order Taylor polynomials.

Supposing that function $f:\R\rightarrow\R$ has derivatives of all orders at point $p$, the \textit{Taylor series} $T_p(f)$ of  $f$ about point $p\in\R$ is defined by  $$T_p(f)(x) \defeq \diss_{j=0}^\infty \frac{f^{(j)}(p)}{j!}(x-p)^j,$$ where $f^{(j)}(p)$ denotes the $j$th derivative of $f$ evaluated at point $p$. When $f(x)=T_p(f)(x)$ for all $x\in \R$, we say that $f$ is \textit{real analytic}. The degree $n$ \textit{Taylor polynomial} is defined $$T_{p,n}(f)(x) \defeq \diss_{j=0}^n\frac{f^{(j)}(p)}{j!}(x-p)^j $$ and for real analytic functions approximates $f$ away from $p$ as $n$ grows. 
 The convergence is  uniform: for any bounded interval $[a,b]\subset\R$ and precision specification $\e$, there is a number $N_\e>0$ for which \begin{equation}\label{eq:uniform_convergence_taylor}\max_{x\in [a,b]} \left| T_{p,n}(f)(x) - f(x) \right| <\e \end{equation} as long as  $n\geq N_\e$. Thus knowledge of derivatives at a single point provides both quantified local  ($o(h^n)$) and global (uniform convergence on arbitrarily large bounded intervals) information. We import this perspective into a learning scheme, which emphasizes using data  concentrated about a point, instead of all data, as a traditional empirical risk minimization learning algorithm otherwise would.

\subsection{$\Pc$-Learnability}\label{subsection:P-Learning}
The familiar notion of probably approximately correct (PAC) learnability guarantees generalization performance (precision) of an algorithm for a  hypothesis class $\Hc\subset\Yc^\Xc$ with high enough probability (confidence), provided enough data is fed into a learning algorithm (\cite{uml}, \cite{fml},  and \cite{fds}). Both precision and confidence are arbitrarily specifiable, and the varying dependent parameter is  sample complexity (amount of data required) to satisfy these specifications.  In this framework, satisfaction  is guaranteed with fixed sample complexity, independent of distribution. 

Formally, for cost function $c_{(\cdot)}:\Hc\rightarrow \R^{\Xc\times\Yc}$,  a hypothesis class $\Hc$ is PAC learnable  if there is algorithm $\hy:(\Xc\times\Yc)^\omega\rightarrow\Hc$  and sample complexity $\mu:(0,1)^2\rightarrow\N$ for which $$\Pb_{(\Xc\times\Yc)^m}\left(\Eb(c_{\hy_{(\cdot)}}) - \inf_{\tilde{y}\in \Hc}\Eb(c_{\tilde{y}})>\e\right) < \d$$ whenever $m\geq \mu(\e,\d)$, for all  measures $\Pb_{\Xc\times\Yc}$ on $\Xc\times \Yc$. 
This formal notion of learning is particularly suitable in the context of classification $\ytil:\Xc\rightarrow \{0,1\}$ with cost function $c_{\ytil}(x) \defeq \one_{y\neq \ytil(x)}$. In this case, the expected cost is bounded above by $1$, and  it is reasonable to seek convergence uniform over measures. With regression, by contrast, cost may grow without bound $\sup_{x\in \Xc}c_{\hy}(x) = \infty$ and as such speed of learning ($\m(\cdot,\cdot)$  from PAC) may very much be measure-dependent.

In any event, often a primary aim in learning is  to ensure that given any measure $\Pb_{\Xc\times\Yc}$,  we can well model \textit{it} with enough data, not: given enough data, we can well model \textit{any} measure. Some tasks just are harder to learn than others.  Moreover, learnability may  be possible only for certain measures.  We collect this observation into a nonuniform notion of learning (\cite{schmidtOverfitting}):
\begin{definition}\label{def:p-learnable}
Let $\Pc$ be a collection of probability measures on $\Xc\times\Yc$. We say that a hypothesis class $\Hc\subset \Yc^\Xc$ is $\Pc$-\textit{learnable} if there is sample complexity $\mu:(0,1)\rightarrow\N$ and algorithm $\hy:(\Xc\times\Yc)^\omega\rightarrow \Hc$ for which $$\Pb_{(\Xc\times\Yc)^m}\left(\Eb(c_{\hy_{(\cdot)}}) - \inf_{\tilde{y}\in\Hc}\Eb(c_{\tilde{y}}) > \e \right)<\d$$ whenever $m>\mu(\e,\d)$ and $\Pb_{\Xc\times\Yc}\in \Pc$.  We say that $\Hc$ is   $\Pb_{\Xc\times\Yc}$-\textit{learnable} if $\Hc$ is $\{\Pb_{\Xc\times\Yc}\}$-learnable, and that $\Hc$ is $\Pc$-\textit{nonuniform learnable} if $\Hc$ is  $\{\Pb_{\Xc\times\Yc}\}$-learnable for each $\Pb_{\Xc\times\Yc}\in \Pc$. 
\end{definition}

Compare this definition with a similar  notion of nonuniform learnability in \cite{uml}.

\begin{remark}
Observe again a difference in quantification from  PAC, where we allow specification of  the probability measure before the sample complexity. This is not an accident: we want a notion of learning which  is rich enough to allow for learning of each task, and still makes sense of generalization, but  leaves open the possibility that some tasks may require more data than others. 
\end{remark}

We will impose some restriction on measures (\cref{def:p-learnable}) for the statement and proof of the main theorem (\cref{theorem:main}).  Specifically, we require that marginal $\Pb_\Xc$ be \textit{subguassian} (\cref{def:subgaussian}), and that conditional  $\Pb_{\Yc|\Xc}$ be deterministic for some subexponential real analytic function (\cref{def:subexp}),  for all $\Pb_{\Xc\times\Yc}\in \Pc$. 

\begin{definition}\label{def:subgaussian} A probability measure $\Pb_\R$ on $\R$ is said to be \textit{subgaussian} if there is $c>0$ for which $$\Pb_\Xc\left(|x|>T\right) \leq e^{-cT^2}$$ for all $T\geq 0$ (\cite{hdp}). 
\end{definition}

The subgaussian  condition translates loosely as ``the measure does not have fat tails.'' Subgaussian measures are fairly common: gaussian measures are subgaussian, as is any measure whose support is bounded (e.g.\ a uniform measure on $[a,b]\subset\R$), and many engineering techniques rely on gaussianity of some underlying process (Kalman filters, e.g.).

Subexponentiality of real-analytic function guarantees that a function may be bounded by an exponential one:
\begin{definition}\label{def:subexp}
A real analytic function $f:\R\rightarrow\R$ is subexponential if, for Taylor expansion $f = \diss_{n=0}^\infty a_n(\cdot)^n$, there is some $K>0$ for which $n!|a_n|\leq K^n$ for all $n\in \N$. 
\end{definition}

With this setup, we proceed to the learning theorem.

\section{Taylor Learning Theorem}\label{sec:taylor_learning}

In this section, we  state, discuss, and prove \cref{theorem:main}, which provides a $\Pc$-nonuniform learnability guarantee for a hypothesis class of polynomials. In \cref{subsec:mainTheorem} we introduce the main theorem, and outline its proof. We then state three lemmas and use them to prove the main result. Finally, in \cref{subsection:lemmaProofs}, we provide  proofs of the lemmas.

\subsection{The Theorem}\label{subsec:mainTheorem}
In what follows, we take the cost function to be $c_{\ytil}(x,y)\defeq \big|y-\ytil(x)\big|.$

\begin{theorem}\label{theorem:main}
Let $\Pc$ be class of  measures $\Pb_{\Xc\times\Yc}$ on $\Xc\times\Yc$ satisfying the following two properties: \begin{enumerate}
	\item The marginal $\Pb_\Xc$ is subgaussian (\cref{def:subgaussian}), and 
	\item the conditional $\Pb_{\Yc|\Xc}$ is deterministic: $\Pb_{\Yc|\Xc}(y = f(x)|x)=1$ for some real-analytic subexponential function $f:\R\rightarrow\R$ (\cref{def:subexp}).  
\end{enumerate} Let  $\Hc=\R[x]$ be collection of polynomial functions. 
Then $\Hc$ is $\Pc$-nonuniform learnable (\cref{def:p-learnable}). 
\end{theorem}

\begin{remark}
By condition 2 in the statement of \cref{theorem:main}, we may write cost as $c_{\tilde{y}}(x)$ to denote $|y-\tilde{y}(x)|$ under supposition that $y=y(x)$. We thereby  express the expectation of cost as  $\Eb(c_{\tilde{y}})= \disg_\Xc c_{\tilde{y}}(x)d\Pb_\Xc(x)$ because $\Pb_{\Yc|\Xc}(y|x)=\one_{y=f(x)}$. 
\end{remark}

We  outline  the proof of \cref{theorem:main} as follows. To bound the expected cost, we split the integral into two components: a tail component consisting of the integral from some arbitrarily large endpoint to infinity (similarly for the negative axis tail), and a body component consisting of the  integral on a compact domain. We then show that each can be made arbitrarily small. We start with the tail and use subgaussianity with exponential boundedness of $y$---which implies that $|y(x)|\leq e^{\eta x}$--- to show that the tail integral  $\disg_T^\infty c_{\ytil}(x)d\Pb_\Xc(x)$  converges, and therefore can be made arbitrarily small with sufficiently large endpoint $T$. Then we are left to deal with the body component  $\disg_{-T}^T c_{\ytil}(x)d\Pb_\Xc(x)$.  Bounding this integral amounts to bounding the integrand: $$\disg_{-T}^T  c_{\ytil}(x)\,d\Pb_\Xc(x) \leq 2T \dissup_{x\in [-T,T]} \big|y(x)-\ytil (x)\big|.$$
The term $|y(x)-\ytil (x)|$ is further bounded by \begin{equation}\label{eq:taylorBound} \diss_{j=0}^N \frac{\big|y^{(j)}(0)-\ytil ^{(j)}(0)\big|}{j!}|x|^j + \left|\diss_{j=N+1}^\infty \frac{y^{(j)}(0)}{j!}x^j \right|, \end{equation} the  sum of differences of Taylor polynomials and the tail of a Taylor series. The tail of the series can be made arbitrarily small by taking enough terms for the Taylor polynomial ($y(x)$ is real anaytic). Up to this point, our argument makes no reference to or use of sampling data. It is in finally bounding the error of derivative approximations $\ytil{(j)}(0)$ that ``learning from data'' occurs.   For this bound we need a collection of lemmas, interesting in their own right.

The first one (\cref{lemma:sampling}) says that we can, with arbitrarily high probability, guarantee sampling an arbitrary number of sampled data points from a positive probability event, provided we sample enough. The next lemma (\cref{lemma:point}) guarantees that we can  make use of \cref{lemma:sampling}: there is a point around which any arbitrary interval has positive probability measure. And the last one (\cref{lemma:derivative_approximation})  ensures that we can arbitrarily well approximate any $n$-th order derivative as long as we take enough function evaluations close enough to the point to be approximated.  Taken together, lemmas \ref{lemma:sampling}, \ref{lemma:point}, and \ref{lemma:derivative_approximation} imply that with arbitrarily high probability we may arbitrarily well approximate any $n$-th order derivative. 

\begin{lemma}\label{lemma:sampling}
Suppose $\Pb_\Xc\big((p-h,p+h)\big)>0$, and let $m\in \N,\ \d >0$. Then there exists $M\in\N$ such that for  $x_1,\ldots,x_M\sim_{iid}\Pb_\Xc$ independently sampled data points, at least $m$ of them $x_{i_1},\ldots,x_{i_m} \in (p-h,p+h)$ with probability at least $1-\d$.
\end{lemma}
\begin{remark} Formally writing this statement as an event, \footnotesize  $$
\Pb_{\Xc^M}\left(\disu_{1\leq i_1\leq\ldots\leq i_m\leq M }\big\{(x_1,\ldots,x_M)\in \Xc^M:\,  (x_{i_1},\ldots,x_{i_m})\in (p-h,p+h)^m\big\}\right) \geq 1- \d.$$ 
\end{remark}
\normalsize 
The next lemma guarantees the general applicability of    \cref{lemma:sampling}. 

\begin{lemma}\label{lemma:point}
Let $\Pb_\Xc$ be a probability measure on $\R$. Then there is a point $p\in \R$ such that $$\Pb\big((p-h,p+h)\big)= \disg_{p-h}^{p+h} d\Pb(x) >0$$ for every $h>0$.
\end{lemma}

\begin{lemma}\label{lemma:derivative_approximation}
Let $f:\R\rightarrow\R$ be a real analytic function.
For every $\e>0$ and $n\in \N$, $f^{(n)}(0)$ may be estimated to arbitrary precision with $n+1$ sampled data points $f(h_0),\ldots, f(h_n)$, i.e.\ $$f^{(n)}(0) = \diss_{j=0}^n a_jf(x_j) + o(h)$$ with $\sup_{i,j}|h_j-h_i| \leq Ch$ for some constant $C$. 
\end{lemma}
 
 In fact, a tighter error term $o(h^p)$ may be achieved by taking at least $k+p$ sampled points $h_0,\ldots,h_{p+k-1}$ (\cite{leveque}, see also  \cite{fornberg}, \cite{fornberg2}, \cite{bowen}).

Having introduced the lemmas, we now prove \cref{theorem:main}. 

\begin{proof}[Proof of \cref{theorem:main}]

Suppose first, without loss of generality, that $0$ is a point  which satisfies \cref{lemma:point}, namely $\Pb\big((-\g,\g)\big)>0$ for each $\g>0$. Fix $\d,\e>0$. We wish to show that there is integer $M_{\d,\e}>0$ such that sampling $x_1,\ldots,x_M\sim_{iid}\Pb_\Xc$ data points guarantees, with probability at least $1-\d$, a bound on error \begin{equation}\label{eq:goal}\disg_{-\infty}^\infty c_{\ytil}(x)d\Pb(x) = \disg_{-\infty}^\infty \left|y(x)-\ytil(x)\right|d\Pb(x) < \e.  \end{equation}

We split the integral in  \eqref{eq:goal} into three components:
$$\disg_{-\infty}^\infty c_{\ytil}(x)d\Pb(x) = \disg_{-\infty}^{-T} c_{\ytil}(x)d\Pb(x) + \disg_{-T}^T c_{\ytil}(x)d\Pb(x) + \disg_{T}^\infty c_{\ytil}(x)d\Pb(x)$$ and bound each. Since the first and third are conceptually identical, we present the argument bounding the third.  By assumption there is some bound $K$ such that $|y^{(j)}(0)| < K^j$ for all $j\geq 0$. Therefore, $$\begin{array}{ll} |y(x)-\ytil(x)| & \leq |y(x)|+|\ytil(x)|\\ &  =\left| \diss_{j=0}^\infty \frac{y^{(j)}(0)}{j!}x^j\right| + |\ytil(x)|\\ & \leq \diss_{j=0}^\infty \frac{|y^{(j)}(0)|}{j!}|x|^j + |\ytil(x)| \\ & \leq \diss_{j=0}^\infty \frac{|Kx|^j}{j!}+ |\ytil(x)| \\ &= e^{|Kx|} +|\ytil(x)|.\end{array}$$ A similar bound can be made for $|\ytil(x)|$. Therefore, we may bound $|c_{\ytil}(x)| < e^{K|x|}$ for all $|x|> T$. As $\Pb_\Xc(x)$ is subgaussian,  we claim that the  integral \begin{equation}\label{eq:improper_1}\disg_T^\infty e^{K|x|}d\Pb_\Xc(x)<\infty\end{equation}  integral converges, so that  $$\disl_{T\rightarrow\infty} \disg_T^\infty e^{K|x|}d\Pb_\Xc(x) = 0,$$  and each tail can be bound by $\e/4$ for sufficiently large $T$.

To show convergence of integral in  $\cref{eq:improper_1}$, without loss of generality suppose that $T=0$ and   observe that $x=\disg_0^xdt = \disg_0^\infty \one_{x\geq t}dt$. We then write \begin{equation}\label{eq:subgauss1} \disg_0^\infty e^{Kx}d\Pb_\Xc(x)  = \disg_0^\infty\disg_0^\infty \one_{e^{Kx}\geq t}dt\,d\Pb_\Xc(x) 
 = \disg_0^\infty \disg_0^\infty \one_{e^{Kx}\geq t}d\Pb_\Xc(x)\, dt 
 = \disg_0^\infty \Pb_\Xc\big(e^{Kx}\geq t\big)  dt 
\end{equation} where the  second equality follows by Tonelli, and the third since $\Eb(\one_E) = \Pb(E)$.

Now fix $p>1$ and $c$ as in \cref{def:subgaussian},  let  $t^* = \max\big\{1,e^{\frac{K^2p}{c}}\big\}$, and observing that  \begin{equation}\label{eq:loglog} -c\left(\frac{\log(t)}{K}\right)^2 = - c\left(\frac{\log(t)}{K^2}\right)\log(t) = -\log\left(t^{c\frac{\log(t)}{K^2}}\right), \end{equation}  we split the last integral in \cref{eq:subgauss1} and bound as  \begin{equation} \label{eq:subgauss2} \disg_0^{t^*} \Pb\big(e^{Kx}\geq t\big)  dt  + \disg_{t^*}^\infty \Pb\big(e^{Kx}\geq t\big)  dt\leq t^* + 2\disg_{t^*}^\infty e^{-c\left(\frac{\log(t)}{K}\right)^2}dt = t^* + 2\disg_{t^*}^\infty \frac{1}{t^{c\frac{\log(t)}{K^2}}}dt. \end{equation}   The inequality holds  by bound $\Pb(E)\leq 1$ and subgaussianity of $\Pb_\Xc$ (\cref{def:subgaussian}), and the equality by \cref{eq:loglog}. Finally, for $t>t^*$, $t^{c\frac{\log(t)}{K^2}} \geq t^{p}$, so the final integral in \cref{eq:subgauss2}  is bounded above as $$\disg_{t^*}^\infty \frac{1}{t^{c\frac{\log(t)}{K^2}}}dt\leq \disg_{t^*}^\infty \frac{1}{t^{p}}dt,$$ which converges by the integral p-test, proving that integral \eqref{eq:improper_1} converges and  indeed \begin{equation}\label{eq:T_epsilon} \disl_{T\rightarrow\infty} \disg_T^\infty e^{Kx}d\Pb_\Xc(x) = 0.\end{equation}

We now bound the  integral \begin{equation}\label{eq:body} \disg_{-T}^Tc_{\ytil}(x)d\Pb_\Xc(x) \leq \disg_{-T}^T  \diss_{j=0}^N \frac{\big|y^{(j)}(0)-\ytil^{(j)}(0)\big|}{j!}|x|^j + \left|\diss_{j=N+1}^\infty \frac{y^{(j)}(0)}{j!}x^j\right| d\Pb_\Xc(x),\end{equation} recalling \cref{eq:taylorBound}, where $T$ is chosen so that the sum of tail integrals is  bounded by $\e/2$ from \eqref{eq:T_epsilon}. We write \cref{eq:body} as $$\underbrace{\disg_{-T}^T  \diss_{j=0}^N \frac{\big|y^{(j)}(0)-\ytil^{(j)}(0)\big|}{j!}|x|^jd\Pb_\Xc(x)}_{I_1(N)} + \underbrace{\disg_{-T}^T \left|\diss_{j=N+1}^\infty \frac{y^{(j)}(0)}{j!}x^j\right| d\Pb_\Xc(x)}_{I_2(N)},$$ and bound each $I_j(N)$ individually. On compact $[-T,T]$ convergence of Taylor series for  $y:\R\rightarrow\R$ is uniform (see \cref{eq:uniform_convergence_taylor}) which in particular implies that  $$\disl_{N\rightarrow\infty}\dissup_{x\in [-T,T]} \left|\diss_{j=N+1}^\infty \frac{y^{(j)}(0)}{j!}x^j\right| = 0,$$ and hence that $\disl_{N\rightarrow\infty}I_2(N) = 0$.  Select $N$ so that $I_2(N)<\e/4$; then we are left with bounding $I_1(N)$.

Up to this point, we have not used the fact that $\ytil=\hy_{\sF{S}}$ is a model constructed from data. Indeed, we have yet to specify how $\ytil$ is constructed. To bound the estimate $I_1(N)$, we cite \cref{lemma:point}, \cref{lemma:sampling}, and \cref{lemma:derivative_approximation}. First, we isolate the terms to bound: $$\begin{array}{ll}
\disg_{-T}^T\diss_{j=0}^N\frac{|y^{(j)}(0)- \ytil^{(j)}(0)|}{j!}|x^j|\, d\Pb(x) & \leq 2T \dissup_{x\in[-T,T]}\diss_{j=0}^N\frac{|y^{(j)}-\ytil^{(j)}(0)|}{j!}|x|^j \\ & \leq 2T^{N+1}\diss_{j=0}^N\frac{|y^{(j)}(0)-\ytil^{(j)}(0)|}{j!}\\
& \leq 2T^{N+1}(N+1)\cdot \dismax_{j=0,\ldots,N}\frac{|y^{(j)}(0)-\ytil^{(j)}(0)|}{j!}, 
\end{array}$$ where we use  $\Pb_\Xc([-T,T])\leq 1$ for the first inequality and suppose without loss of generality that $T\geq 1$ in the second. Thus we must show that bound $$\dismax_{j=0,\ldots,N}\frac{|y^{(j)}(0)-\ytil^{(j)}(0)|}{j!} \leq \tilde{\e}\defeq \frac{\e}{8(N+1)T^{N+1}} $$ is achievable in probability, 
which bound is certainly satisfied if for each $j=0,\ldots,N$, $$|y^{(j)}(0) - \ytil^{(j)}(0)|\leq \tilde{\e}.$$  Indeed, \cref{lemma:derivative_approximation} guarantees that this bound may be achieved with enough data sampled, say $m$ points, in an interval $(-\d_j(\tilde{\e}),\d_j(\tilde{\e}))$, \cref{lemma:point} guarantees the existence of point with positive probability for every  interval about the point (we suppose without loss of generality that $0$ is such a point),  and \cref{lemma:sampling} guarantees the existence of $M>0$ for which $m$ points $x_{i_1},\ldots,x_{i_m}$ of $x_1,\ldots,x_M$ are with probability at least $1-\d$  in $(-\d_j(\tilde{\e}),\d_j(\tilde{\e}))$. \end{proof}

\begin{remark}
The restriction on class of functions (subexponential) and marginal distribution (subgaussian) are not particularly strong, but in nearly every engineering application the conditional probability  $\Pb_{\Yc|\Xc}$ will not be deterministic ($\Pb_{\Yc|\Xc}(y_i=y(x_i)|x_i)\neq 1$)  because the underlying process is inherently non-deterministic or---what amounts to much of the same---because there is noise in measurement.  In such case, any realization of noise (as a random variable) will (most likely) be everywhere discontinuous, and therefore not smooth, and very much therefore not real analytic. 
 
 This observation does not automatically render \cref{theorem:main}  inapplicable to engineering problems. A class of engineering's accomplishments is the development of sensors: with one we may directly observe e.g.\ position, and with another directly observe acceleration. To the extent that mathematics can make requests on the engineering community, we here have another instance: more sensors to directly measure higher order derivatives may directly inform and support our algorithms. 
\end{remark}

\subsection{Proofs of Lemmas}\label{subsection:lemmaProofs}
We start by proving the lemmas and end with the proof of \cref{theorem:main}.

\begin{proof}[Proof of \cref{lemma:sampling}]
Set $\g \defeq \Pb_\Xc\big((p-h,p+h)\big) $ and let $E\subset \R^M$ be the event that at most $m-1$ elements $\{x_{j_1},\ldots,x_{j_{m-1}}\}$ are contained in $(p-h,p+h)$: $$E\defeq \big\{q\in \R^M:\, \mbox{at most }\, m-1\; \mbox{factors}\; q_{i_1},\ldots,q_{i_{m-1}} \in (p-h,p+h)\big\},$$ and $E_k$ the event that exactly $k$ elements $\{x_{j_1},\ldots, x_{j_k}\}\subset (p-h,p+h)$. Let $\Pb_{\Xc^j}$  denote the product (independent) probability measure (induced by $\Pb_\Xc$) on $\R^j$ and observe that since $E=\discats_{j=0}^{m-1}E_j$ is a disjoint union, we have $$\begin{array}{ll} \Pb_{\Xc^M}(E) & = \diss_{j=0}^{m-1}\Pb_{\Xc^M}(E_j) \\
& = \diss_{j=0}^{m-1}\begin{pmatrix} M \\ j
\end{pmatrix} \left(\Pb_{\Xc}\left(\big\{x\notin (p-h,p+h)\big\}\right)\right)^{M-j}\left(\Pb_\Xc\left(\big\{x\in(p-h,p+h)\big\}\right)\right)^j \\
& = \diss_{j=0}^{m-1}\begin{pmatrix} M\\ j
\end{pmatrix} (1-\g)^{M-j}\g^j \\
& \leq m K (1-\g)^{M-m}\cdot 1 \\ 
& \leq cmM^m(1-\g)^M \xrightarrow{M\rightarrow\infty}0 ,
\end{array}$$ where $K=\max\left\{\begin{pmatrix} M \\ j \end{pmatrix}:\, j =0,\ldots, m-1\right\}\leq cM^m$ (which max will in general be realized at $m-1$ when $M>>m$). \end{proof}

\begin{proof}[Proof of \cref{lemma:point}]
By continuity of measure (\cite{pwm}), there is some bounded interval $I=[a,b]\subset \R$ with $\Pb(I) > 0$. We argue inductively by selecting descending sequence of positive measure sets: at stage $j=0$, $I_j=I$, with positive measure.  At stage $j+1$, we have interval $I_j=[a_j,b_j]$, and   let $A_j=\left[a_j,\frac{b_j-a_j}{2}\right]$ and $B_j=\left[\frac{b_j-a_j}{2},b_j\right]$. Since $\Pb(I_j) >0$, at least one of  $\Pb(A_j) >0$  or $\Pb(B_j) >0$, and set $I_{j+1}$ to be either interval of positive measure. Observe that $I_{j+1}\subset I_j$ and that we therefore have decreasing sequence $$I_0\supset I_1\supset \cdots \supset I_j\supset I_{j+1}$$ of nonempty closed sets. By Cantor's Intersection Theorem (\cite{follandanalysis}), there is at least one point  $$p\in \disi_{j=0}^\infty I_j.$$ Since any interval $(p-h,p+h)$ about $p$, for $h>0$, contains at least one $I_j$ (in fact, infinitely many), we conclude that $\Pb\big((p-h,p+h)\big)>0$.\end{proof}

\bibliographystyle{ieeetr}
\bibliography{phdref}

\end{document}